\documentclass[letterpaper, 10 pt, conference]{ieeeconf} 

\IEEEoverridecommandlockouts 

\usepackage{graphicx}

\title{\LARGE \bf
The Social Triad model of Human-Robot Interaction*
}

\author{David Cameron$^{1}$ and Emily Collins$^{2}$ and Stevienna de Saille$^{3}$ and James Law$^{4}$
\thanks{*This work was funded by the UKRI projects EP/V00784X/1 Trustworthy Autonomous Systems Hub}
\thanks{$^{1}$David Cameron is with the Information School, University of Sheffield, UK
 {\tt\small d.s.cameron@sheffield.ac.uk}}%
\thanks{$^{2}$Emily Collins is with the Institute for Experiential Robotics, Northeastern University, Mass, USA
 {\tt\small e.collins@northeastern.edu}}%
\thanks{$^{3}$Stevienna de Saille is with the Department for Sociologial Studies, University of Sheffield, UK
 {\tt\small s.desaille@sheffield.ac.uk}}%
 \thanks{$^{4}$James Law is with the Department of Computer Science, University of Sheffield, UK
 {\tt\small j.law@sheffield.ac.uk}}%
}

\begin{document}

\maketitle
\thispagestyle{empty}
\pagestyle{empty}

\begin{abstract}
Despite the increasing interest in trust in human-robot interaction (HRI), there is still relatively little exploration of trust as a social construct in HRI. We propose that integration of useful models of human-human trust from psychology, highlight a potentially overlooked aspect of trust in HRI: a robot’s apparent trustworthiness may indirectly relate to the user’s relationship with, and opinion of, the individual or organisation deploying the robot. Our Social Triad for HRI model (User, Robot, Deployer), identifies areas for consideration in co-creating trustworthy robotics.

\end{abstract}

\section{INTRODUCTION}

Trust is increasingly recognised as being an important aspect of successful Human-Robot Interaction (HRI). Substantial strategic funding has been invested (e.g., Trustworthy Autonomous Systems, www.tas.ac.uk) to address the topic, from direct study of HRI through to understanding the wider socio-technical contexts for creating trustworthy robotics. This short paper proposes a means by which the wider contexts for trustworthy robotics may shape trust during HRI and potentially the outcomes of HRI studies; specifically, we consider how wider relationships in an HRI study could affect trust - as a social construct - towards robots.

Despite being relatively well-trodden ground in psychological literature that evaluations of others (and trust towards others) can be characterised as a social construct having both cognitive and affective components \cite{weiss2021trust,mcallister1995affect,fiske2007universal}, exploration of trust as a social construct is still a relatively new area of study for HRI. While established models of trust such as the Human, Robot, and Environment model \cite{hancock2011meta,schaefer2016meta,hancock2021evolving} cover a wide range of factors, people's experiences of trust towards robots - and even social robots at that - has been largely explored solely in terms of cognitive factors such as beliefs of capability or reliability \cite{ruff2002human,de2011adaptive,desai2013impact,salem2015would,hancock2011can} rather than affective factors. 

\subsection{Social Robotics}

At its heart, the field of social robotics takes two key directions: simulating social processes in robotic agents and studying people's social experiences and interaction with robotics \cite{sheridan2020review}. In terms of trust, this could include presenting various social-like behaviours in robots to shape interaction experience \cite{law2021trust}, and exploring the potential for social models (e.g., \cite{weiss2021trust,mcallister1995affect,fiske2007universal}) to further understand and explain trust in HRI.

Ahead of the development of any structured model of trust as a social construct for HRI, a range of studies have nonetheless demonstrated user trust can be influenced through robots' simulated social or affective interactions. Examples include using rhetorical persuasion \cite{lee2019robotic}, using gesture and expression \cite{salem2013err}, taking blame \cite{kaniarasu2014effects} or offering apologies for errors \cite{cameron2021effect,koxtrust2022}, and making promises to change behaviours \cite{robinette2015timing}. Emerging social models, drawing from such examples, argue that trust as a social construct, as seen in human-human interaction \cite{weiss2021trust,fiske2007universal}, has relevance in HRI \cite{cameron2021effect, malle2021multidimensional}. People's trust towards a social robot may not be just based on factors such as reliability or capability (the `competence' dimension \cite{fiske2007universal}), but also along the `warmth' dimension based on factors such as apparent integrity or benevolence.

A substantial challenge in integrating social models of trust into HRI research is clearly understanding people's views of a robot as being a social entity and its capacities for such social aspects. Evaluations of early measures of trust as a social construct in HRI highlight a sizable percentage of individuals declaring that concepts such as `intention' and `benevolence' to not apply to robots \cite{chita-tegmark2021}. Yet, the social strategies identified above appear to affect people's trust towards robots; indeed, one study finds apologies promote intentions to use while adversely affecting perceptions of a robot's capability \cite{cameron2021effect}. 

This apparent contradiction still needs resolving; it is unclear how people who do not believe a robot has capacity for (e.g.) `warmth' based social interaction and experiences may still be affected in their trust judgements towards a robot exhibiting simulations of such interactions. In brief, are users in some way being deceived or is there an alternative possibility?

\subsection{Can a social robot be trusted?}

The issue of deception in social robotics is still hotly contested \cite{emilydeception}, though one perspective \cite{amandadeception} offers some points towards clarity in understanding trust as a social construct in HRI. 

As a starting point, the paper argues robots have no intention; despite their apparent independence and agency, their behaviours are reflections of the limitations set out by programmers \cite{amandadeception}. That said, the paper recognises that it is possible to deceive without intention, but again argues the deception comes not from the robot per se; instead the deception arises from the design and programming of the robot, the circumstances to create the deception, and the users participation in the deception \cite{amandadeception}. Similar arguments liken social robots to that of puppets, in which people interact with the object \textit{as if} they are social agents \cite{Clark2022social}, actively participating in the deception. 

This stance, while ultimately arguing against social robots as being seen as social agents, does nonetheless inform how trust as a social construct has relevance to HRI. Interacting with robots \textit{as if} they are social agents may enable the effects from social strategies to shape trust (outlined earlier) to be seen. Concurrently, knowledge or beliefs that the robot is a depiction of a social agent rather than having the genuine capacity of a social agent, permit honest responses that attributes or behaviours such as integrity or benevolence are not applicable to robots. We pose that trust as a social construct (e.g., understanding its benevolence) is not directed towards the robot itself but towards it as an extension of the person behind the robot. The classic model of a dyad in social robotics HRI is perhaps better examined as a triad.

\section{The Social Triad of HRI}

The robot as puppet analogy \cite{Clark2022social} highlights the role of the \textit{authority} behind the \textit{character} in controlling its interaction with the participating \textit{audience}. The character serves as both as a participant in the interaction and as a mediator for the interaction between audience and authority. We draw from this to develop the Social Triad of HRI, specifically modifying the nature of the authority to better accommodate experienced HRI. In the analogy the authority directly controls the passive puppet; in HRI the authority (we have termed elsewhere the `Deployer' \cite{Cameron2021triad}) may have no direct control over the robot but rather has assumed responsibility for the deployment of the robot and creation of the HRI scenario. 

The Deployer may have programmed the robot's interactions and behaviour, although this is not a necessary requirement. The Deployer may be present during the HRI scenario and active or passive in the interaction between the remaining agents: User and Robot, although again this is not entirely necessary. The Deployer is the individual(s) the User believes has responsibility for the HRI scenario. Concrete examples of a Deployer would include the researcher conducting an HRI experiment or the manager bringing a robot into the workplace. Interactions between the User and the Robot take place \textit{within} the human-human interaction social context (between User and Deployer), seemingly peripheral to the HRI scenario underway.

Within the Social Triad of HRI, trust as a social construct relates to the the User's opinions of the Deployer and potentially of the Robot as an extension, or realisation, of the Deployer's intentions. A Robot may behave with apparent integrity or benevolence, without itself being capable of either. Users may be wholly deceived, actively participate in the deception by responding \textit{as if} it is a social agent, and/or evaluate the Deployer based on the Robot's behaviour. Trust as a social construct as experienced by the User towards the Robot is bound by limits of the User's trust towards the Deployer.

\begin{figure}[hhh]
    \centering
    \includegraphics[width=\columnwidth]{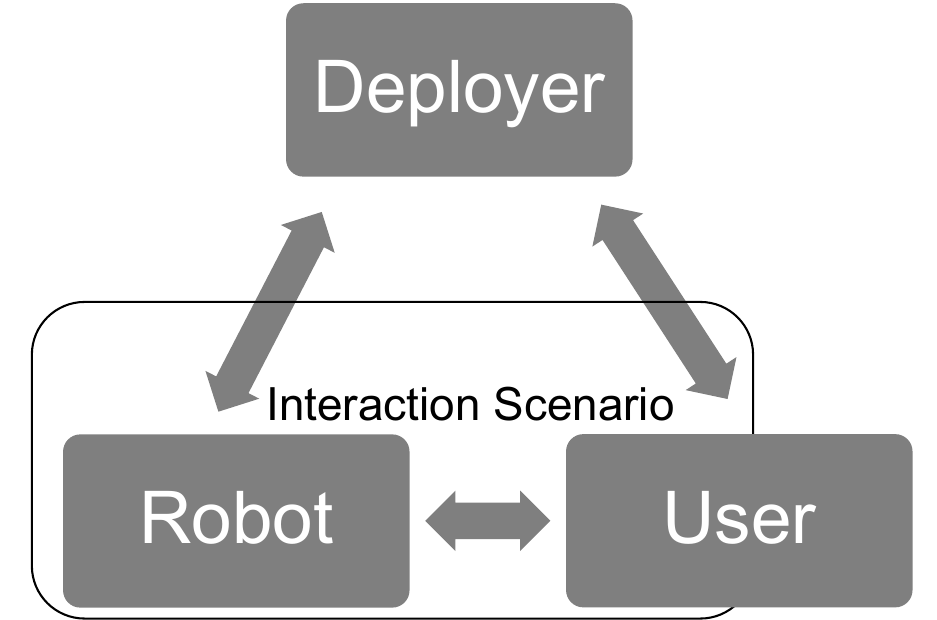}
    \caption{Interaction pathways recognising the role of the Deployer as both external to, and influential on, the HRI scenario. While a robot is contained, the User may enter and exit, the interaction scenario.}
    \label{fig:my_label}
\end{figure}

\subsection{Interactions in the Triad}

The inclusion of a third agent (the Deployer) in HRI adds interaction pathways and relationships to consider. We briefly outline each of the six interaction pathways.

1. HRI is traditionally recognised as just that - the, often reciprocal, human-robot interaction. A User may influence a Robot's behaviour through their own behaviour and communication, either directly controlling or otherwise affecting the robot. 

2. In return, a Robot may shape User trust through its own behaviour, including social-like behaviours. Specifically, User trust may be affected by the robot's appearance and actions \cite{hancock2011meta} and use of simulated social strategies to regulate trust \cite{cameron2021effect,geiskkovitchchildren}.

3. The interaction from Deployer to Robot includes the specification of robot behaviours (e.g., through Wizard of Oz control; specifying goals, or use of architectures for generating behaviour) and specifying the contexts for which the robot may be used, ultimately setting the bounds for a robot's interaction with the User \cite{cameron2015framing}.

4. In return, the Robot provides the Deployer information on the interaction, passively through the behaviour or actively through recorded metrics of the interaction, potentially of both Robot and User.

5. Just as the Deployer determines the bounds of a Robot's behaviour (interaction 3), their formation of the HRI scenario directs the bounds of User behaviour. Existing or emerging relationships between Deployer and User may shape a User's approach to HRI, as would any communication from the Deployer (either directly or via robot behaviour).

6. In return, the User either passively or actively provides feedback to the Deployer on the scenario (in research, this may be from observation, questionnaires etc.; in industry, performance appraisals etc.). Where a User may ostensibly make evaluations of the Robot and HRI scenario, these may reflect their views towards the Deployer's behaviours and intentions; i.e. mistrust towards a Deployer expressed as mistrust towards the Robot.

\section{Co-Creation for Trust}

In sum, where Users place their trust in Robots, they indirectly place their trust in Deployers (cf \cite{Clark2022social}). Simulated social communication from a Robot to engender trust indirectly asks the User to trust the Deployer (i.e. a Robot suggesting its benevolence or it having integrity or showing contrition is a product of the Deployer intending for the User to think of them \textit{both} in these terms). As with deceit not coming from the robot, but as a collaborative effort between the user and those behind the robot and interaction circumstances\cite{amandadeception}, user trust as a social construct is not necessarily \textit{towards} the robot but \textit{around} the robot.

In controlled, laboratory (often university) research, trust towards the Deployer may almost be taken for granted through solemn assurances that the User will come to no harm (physical or psychological); the Robot is programmed to operate appropriately; the specified HRI scenario complies with necessary safety/ethical regulations; and information gathered from HRI will not be used carelessly or maliciously. As participants in such research, Users also have control over engagement with the scenario and power to disengage with the HRI at will to no detriment. Collectively these factors regarding trust towards the Deployer could present an artificially high cap on the User's potential trust towards the Deployer's agent: the Robot. Users may be willing to actively engage with the - as such - deception of social strategies from a robot affecting trust, even while recognising these are not genuinely the robot's. 

The above circumstances in laboratory research may offer little ecological validity, especially in attempts to import to circumstances where there is little trust towards the Deployer and/or little opportunity for them to demonstrate trustworthiness (e.g., introducing a social robot to the workforce). Without a user willing to engage in the deception, a robot's apologies or expressions of benevolence could ring hollow and offer no meaningful indication of the Deployer's intentions. It becomes incumbent on the Deployer to authentically obtain trust before a robot can simulate obtaining trust.

Co-creation as dialogue with the User may be a mechanism by which a Deployer can themselves earn trust, in turn scaffolding trust as a social construct for HRI. Where co-creation of specific robotic systems can require extensive resources and investment, co-creation of the interaction scenarios may meaningfully allow potential Users to receive assurances and control over interaction (akin to the above laboratory research) at comparatively low investment. User control over the types of simulated social behaviours and communications they would be willing to accept - as such, the deceptions they are happy to engage in - may raise their footing from individuals with HRI put upon them to that equal of the Deployer through shared responsibility for the creation of the scenario.

\section{Summary}
Trust research in HRI typically considers just the interactions between user and robot. We propose that while this offers insight into the users' experiences, it omits the important context of relevant human-human interactions that enable or create the HRI scenario. Examining user trust towards the individual(s) that a user believes responsible for the HRI scenario (we have termed Deployer) may further understanding of trust in HRI and inform effective and appropriate means to build trust. Co-creation as a method could be a useful means to explore how changes in relationships between the User and Deployer (namely empowering Users and building trust \textit{between} User and Deployer) may shape trust towards robotics in HRI.

\addtolength{\textheight}{-18cm} 

\end{document}